# METAHEURISTICS IN FLOOD DISASTER MANAGEMENT AND RISK ASSESSMENT

**Presented at**

8th National conference on Information Technology Education (NCITE 2010)
October 20-23, 2010 in Boracay, Philippines

UP ICE Centennial Conference Harmonizing Infrastructure with the Environment
November 12, 2010 in Manila, Philippines

**Vena Pearl Boñgolan[a], Florencio C. Ballesteros[b], Jr., Joyce Anne M. Banting[a], Aina Marie Q. Olaes[a], Charlymagne R. Aquino[b]**

a)Scientific Computing Laboratory, Department of Computer Science
College of Engineering
University of the Philippines Diliman
bongolan@dcs.upd.edu.ph
+639261349191

b) Environmental Engineering Program, College of Engineering
University of the Philippines Diliman

## ABSTRACT

A conceptual area is divided into units or *barangays*, each was allowed to evolve under a physical constraint. A risk assessment method is then used to identify the flood risk in each community using the following risk factors: the area's urbanized area ratio, literacy rate, mortality rate, poverty incidence, radio/TV penetration, and state of structural and non-structural measures. Vulnerability is defined as a weighted-sum of these components. A ‘penalty’ was imposed for reduced vulnerability. Optimization comparison was done with MatLab’s Genetic Algorithms and Simulated Annealing; Results showed ‘extreme’ solutions and realistic designs, for simulated annealing and genetic algorithm, respectively.

## INTRODUCTION

Disaster Risk Management (DRM) at the local, regional, and global scale continues to generate great research interest of a complex, multidisciplinary nature, involving the interplay of scientific, social, economic, and political dimensions. Driven by the series of disasters of increasing frequency and magnitude, DRM meaning and context has evolved into an internationally accepted definition: a systemic approach to identifying, assessing and reducing risk of all kinds associated with hazards and human activities with identified operational and practical disaster risk reduction initiatives. These initiatives have been clarified by the international community through UN’s 2005 World Conference on Disaster Reduction in Kobe, Japan and accepted as the DRR framework, known as the Hyogo Framework of Action [1].

The ultimate objective of all DRM initiatives remains simple: reduce the loss of lives and property, and improve the capacity of communities to cope with disasters. The 2005 Hyogo Framework of Action (HFA) has been used to review UN member states’ respective DRM initiatives. HFA outlines five (5) priorities for implementation: (1) Ensure that disaster risk reduction is a national and a local priority with a strong institutional basis for implementation, (2) Identify, assess and monitor disaster risks and enhance early warning, (3) Use knowledge, innovation and education to build a culture of safety and resilience at all levels, (4) Reduce the underlying risk factors, and (5) Strengthen disaster preparedness for effective response at all levels. This paper, however small, contributes to the fulfillment of HFA's priority action no. 4 that seeks to reduce the underlying (disaster) risk factors.

Adopting flood as the specific natural hazard, this study compares two (2) mathematical methods, Genetic Algorithm and Simulated Annealing, to identify optimal characteristics of a hypothetical city that would lessen flood hazard risks resulting to reduction in costs, i.e. damage of properties and use of resources. Although the results and analysis presented in this study is not intended an existing city, the mathematical program developed in this research may be useful for city planners and urban developers in projected scenarios and plans of a disaster resilient city.

Research and similar studies have been undertaken for the past 10 years. In early 2008, the IBM research laboratories in the U.S and India had launched a disaster management technology/software to help relevant government and non-government agencies and offices model and manage natural disasters, i.e. wildfires, floods, and diseases. Known as Stochastic Optimization Models for Natural Disasters, IBM created a set of intellectual properties and software assets that can be employed to gauge and improve levels of preparedness to tackle unforeseen natural disasters involving most real-world uncertainty factors such as politics, custom and culture making results grow less uncertain and more accurate and acceptable [2].

## BACKGROUND

Genetic Algorithm (GA) is used because of the this case's high dimensionality that is similar to a multi-objective urban planning problem tackled by Balling et al. in [3] where direct methods are intractable, if not impossible to use.

GA, copies the natural evolution, i.e., inheritance, mutation, selection and crossover, has been widely used to solve engineering optimization problems. Adapting concept of dog and/or cattle breeding, the method takes champion males and females to breed, in the hope of producing desirable 'traits' in the offspring. Here, a 'champion' city has the desirable trait of a low vulnerability to flooding. The 'chromosomes' for our 'champion' cities, and the GA will then take care of 'breeding' the cities, looking for the one with lowest vulnerability and cost.

Another widely used metaheuristic that mimics metallurgy's annealing process– that of slow cooling to provide atoms more chances of finding configurations with lower internal energy in order to increase crystal sizes and reduce the defects - the Simulated Annealing (SA) in this study, slowly reduces the extent of the search by looking for 'nearby' points for lower energy-solutions.

This paper uses the same fitness and cost functions to be able to compare designs or solutions from GA and SA, integrating previous research done in [4] and [5] using flood as the natural hazard. All disaster management plans begins with the identification of risk (UN/ISDR, 2004), see [6].

Risk (flood) is an indicator of how prone a specific area is for a natural hazard to turn into a disaster; it is a function of three factors defined by this equation:

Eqn. 1: Flood Risk = Hazard x Vulnerability x Exposure, where

Vulnerability is expressed in a linear, very predictable function as:

Eqn. 2: Vulnerability = (A)Urbanization + (B)Literacy Rate + (C)Mortality Rate + 2* (D)Poverty + (E)Radio/TV + (F)Non-structural measures + (G)Structural measures

Vulnerability is at its minimum when values of each component are zero (0). Weight of each components were assumed and pegged at one (1), while poverty was given twice (2) the weight as suggested by a previous earthquake vulnerability studies by Lo and Oreta [5] posits that poorer communities are more vulnerable to disasters. Studies may be conducted separately to fine-tune values of different components.

Dynamics were introduced into the system by imposing a ‘penalty’ or cost function. Consider poverty: A low poverty area is desirable, as it is deemed less vulnerable to flooding and other hazards. It would have a vulnerability of zero. We now impose a ‘penalty’, the cost of getting out of vulnerability, which might also be interpreted as the cost of the solution to poverty.

Whenever natural hazards are concerned, little can be done on the hazard probability of an area for a specific disaster as it is predominantly defined by geographic location. This is also true for exposure, in the Philippines, a disaster-prone area, i.e. flood prone due to rainfall hazards, is geography-based. The reason why most disaster management plans, if not all, focus on modifying the vulnerability index to minimize risk. This study adopted this approach.

## METHODOLOGY

**Defining the domain and weights.** The hypothetical study area is divided into several computational units also defined as the political boundaries, e.g., *barangays* (communities or neighborhoods in a city). A 6 by 6 grid is used to represent the city and its 36 areas/ barangays. Please refer to Figure 1.

First, we define the physical area. The numbers inside each cell represent physical properties of the area that might make it more or less vulnerable to flooding; each cells represent each barangay. It will also be used to multiply the area’s vulnerability. Here, we may imagine a river running diagonally down the grid, from right to left. Hence, the barangays on the right diagonal lie on a flood plain, and will have a factor of two (to double the vulnerability). We assume the terrain rises a bit above the flood plain on either side of the diagonal. Thus, it will have a value of 1 (will not affect vulnerability). Those barangays farthest from the floodplain, assumed to be on higher ground, will have a factor of ½, to indicate lower vulnerability.

**Figure 1.**
**6x6 grid representing a city**

| | | | | | |
|---|---|---|---|---|---|
| ½ | ½ | ½ | ½ | 1 | 2 |
| ½ | ½ | ½ | 1 | 2 | 1 |
| ½ | ½ | 1 | 2 | 1 | ½ |
| ½ | 1 | 2 | 1 | ½ | ½ |
| 1 | 2 | 1 | ½ | ½ | ½ |
| 2 | 1 | ½ | ½ | ½ | ½ |

Each unit, *barangay*, is considered an organism, with its own set of chromosomes, whose values were initially assigned. The values of each chromosome give the design of the city and will be used to produce the designs in the next iterations. In the final iteration, the optimal set of chromosomes will be interpreted.

**Setting component cost/penalty values.** The objective function for both GA (usually called ‘fitness’ function in GA applications) and SA will have the following components, as stated in the vulnerability value expressed in Eqn. 2:

(A)Urbanized area ratio – Non-urbanized areas are devoid of infrastructure and residents. The less people and infrastructure, the lower the vulnerability to flooding.

11 = Highly urbanized
10 = Moderately urbanized
01 = A little urbanized
00 = Not urbanized

(B)Literacy Rate – This requires that communities understand warning signs from the government and follow instructions from emergency responders. It has implications on the over-all community preparedness to disasters. A high illiteracy rate equals vulnerability to flooding.

11 = more than 75% are illiterate
10 = 50 – 75% are illiterate
01 = 25 – 50% are illiterate
00 = 0 – 25% are illiterate

(C)Mortality Rate – The mortality rate is taken as an indicator of the general health of the population. Although it is closely associated with poverty, it is treated as separate variable as reflection of separate data source, different government agencies, health and poverty data from Department of Health (DOH) and Department of Social Welfare and Development (DSWD), respectively. High mortality rate also mean high vulnerability.

11 = High mortality rate
10 = Average mortality rate
01 = Below average mortality rate
00 = Low mortality rate

(D)Poverty (incidence) – Poor members of community are more vulnerable to disaster; both immediate effect and the aftermath of flood. While floods tend to affect wide areas and all socio-economic classes, it is expected that members with more resources to have a shorter recovery time and have more options that include moving to other non-flooded areas. High poverty incidence (in the area) increases its vulnerability to flooding.

11 = more than 75% in class D
10 = 50 – 75% in class D
01 = 25 – 50% in class D
00 = 0 – 25% in class D

(E)TV / Radio (penetration rate) – Access to communication devices means timely receipt of broadcast news and warning. Mass media is a critical factor in disaster preparedness. Less penetration rate increases vulnerability to flooding.

11 = less than 25% penetration rate
10 = 25 – 50% penetration rate
01 = 50 – 75% penetration rate
00 = 75 – 100% penetration rate

(F)State of non-structural measures – This pertains to legislated laws and, more importantly, compliance to laws that were meant to minimize flood risks, i.e. easements on waterways, cleaning drainages, litter prevention, etc. High incidence of implementation of laws in an area means low vulnerability to flooding.

11 = no non-structural measure
10 = existing with poor implementation/compliance
01 = existing with average implementation/compliance
00 = existing with good implementation/compliance

(G)State of structural measures – Infrastructures built to prevent and/or control floods, i.e. drainage, floodgates, pumping stations, etc. More and better-maintained infrastructures mean less vulnerable to flooding.

11 = no structural measure
10 = existing structural measure in poor condition
01 = existing structural measure in average condition
00 = existing structural measure in good condition

These variables were used as costs and input to vulnerability function that will be minimized using both GA and SA methods.

As stated in Eqn. 2 in the previous section, vulnerability, $V_i$, is expressed simply as a linear sum of the numerical values of the "chromosomes", multiplying the poverty "chromosome" by two. The Lo and Oreta study [5] cited poverty prominently, hence the double weight.

$$V_i = S_i*(X_{iUrbanized} + X_{iLiteracy} + X_{iMortality} + 2*X_{iPoverty} + X_{iTvRadio}+X_{iNonStructural} + X_{iStructural})$$

The cost or penalty function counter-acts the vulnerability value. The chromosomes are modeled as binary numbers, with "00" (zero) being smallest, "11" (three) highest. The penalty function is developed by first taking the three's complement of vulnerability. Thus, the lowest vulnerability will have the highest cost, and vice versa. The complement is then entered into a function, which may be exponential (for 'expensive' items like poverty alleviation); quadratic (for relatively expensive items like improving health), or inexpensive, like formulating laws to protect against flooding. These assessments will vary from place to place, depending on the relative costs of the activity or solution (penalty).

Exponential growth for 'expensive' activities like urbanization, poverty alleviation and building structural measures were assumed. All other chromosomes were assumed to have linear penalties, except the mortality variable, which was assumed to be quadratic. These cost functions are hypotheses, and could be interpolated from data, when available. This could be the object of future research.

$$C_i = (\exp(3-X_{iUrbanized}) + 3 - X_{iLiteracy} + (3 - X_{iMortality})^2 + \exp(3-X_{iPoverty}) + 3- X_{iTvRadio} + 3 - X_{iNonStructural} + \exp(3-X_{iStructural})) / S_i$$

A single-objective Simulated Annealing (SA) was used; the cost function had to be scaled to give roughly the same weight as vulnerability:

$$C_i =(e^{3- XiUrbanized} + 3 - X_{iLiteracy} + (3 - X_{iMortality})^2 + e^{3- XiPoverty} + 3- X_{iTvRadio} + 3 - X_{iNonStructural} + e^{3- XiStructural} )*3.26/S_i$$

Together with the equations stated above, the chromosomes (values) were inputted into MatLab's genetic algorithms toolbox and that of simulated annealing.

## RESULTS AND DISCUSSIONS

GA and SA program iteration runs produced the following optimal arrangements for the 36 barangays, broken-down by their 'traits' or chromosomes as:

**Genetic Algorithm Results:**

1. It allowed 2 poor barangays along the diagonal, flood prone area. This gives us insight on designing housing for flood plains which is a special concern in the Philippines with high poverty incidence;
2. It assumes 9 units to be compliant to rules and regulations and placed worst offenders away from floodplains; and
3. Iteration runs consumed two and half days.

| | | | | | |
|---|---|---|---|---|---|
| 10 | 10 | 11 | 1 | 11 | 1 |
| 0 | 0 | 0 | 1 | 1 | 1 |
| 10 | 0 | 11 | 10 | 10 | 1 |
| 1 | 11 | 0 | 10 | 0 | 1 |
| 10 | 1 | 0 | 1 | 0 | 11 |
| 10 | 11 | 11 | 11 | 10 | 1 |

**Urbanization:** The city plan placed highly urbanized areas, 11, on the flood plain (the main diagonal).

| | | | | | |
|---|---|---|---|---|---|
| 1 | 0 | 10 | 1 | 0 | 0 |
| 11 | 1 | 10 | 1 | 0 | 10 |
| 1 | 0 | 1 | 0 | 10 | 1 |
| 0 | 10 | 11 | 1 | 11 | 10 |
| 11 | 0 | 1 | 10 | 11 | 10 |
| 11 | 10 | 11 | 0 | 11 | 10 |

**Literacy:** There are two 'illiterate' areas, 11, on the floodplain; most were situated away from it.

| | | | | | |
|---|---|---|---|---|---|
| 10 | 0 | 1 | 11 | 10 | 1 |
| 10 | 11 | 11 | 11 | 0 | 10 |
| 1 | 10 | 11 | 1 | 10 | 10 |
| 10 | 0 | 1 | 10 | 0 | 10 |
| 0 | 10 | 11 | 10 | 11 | 10 |
| 11 | 11 | 11 | 10 | 1 | 11 |

**Mortality:** Most cells with highest mortalities, 11, were situated away from the flood plain, except for one.

| | | | | | |
|---|---|---|---|---|---|
| 1 | 10 | 1 | 0 | 11 | 10 |
| 0 | 10 | 1 | 0 | 11 | 1 |
| 0 | 11 | 0 | 1 | 11 | 0 |
| 1 | 1 | 10 | 11 | 1 | 1 |
| 0 | 11 | 0 | 0 | 0 | 1 |
| 0 | 10 | 11 | 11 | 11 | 10 |

**Poverty (incidence):** Seven of the nine poor areas, 11, were situated away from the flood plain.

| | | | | | |
|---|---|---|---|---|---|
| 1 | 10 | 11 | 10 | 10 | 1 |
| 10 | 1 | 0 | 0 | 10 | 11 |
| 11 | 10 | 1 | 0 | 10 | 10 |
| 0 | 10 | 10 | 11 | 10 | 0 |
| 0 | 10 | 10 | 11 | 11 | 0 |
| 10 | 10 | 1 | 0 | 0 | 10 |

**Radio/ TV Penetration:** There are no high-risk areas, 11, in the floodplain.

| | | | | | |
|---|---|---|---|---|---|
| 0 | 10 | 0 | 0 | 11 | 0 |
| 0 | 11 | 1 | 10 | 10 | 0 |
| 11 | 0 | 1 | 1 | 0 | 0 |
| 1 | 11 | 10 | 10 | 1 | 11 |
| 11 | 1 | 10 | 10 | 11 | 1 |
| 1 | 10 | 10 | 1 | 10 | 10 |

**State of non-structural measures:** On laws and compliance, no high-risk areas, 11, in the floodplain.

| | | | | | |
|---|---|---|---|---|---|
| 11 | 1 | 1 | 1 | 10 | 10 |
| 0 | 10 | 10 | 1 | 0 | 10 |
| 11 | 1 | 1 | 10 | 11 | 0 |
| 1 | 0 | 1 | 0 | 11 | 10 |
| 11 | 10 | 10 | 11 | 11 | 10 |
| 0 | 0 | 1 | 10 | 0 | 0 |

**State of structural measures:** No high-risk areas, 11, in the flood plain

**Simulated Annealing Results:**

1. SA searches nearby solutions, whereas GA randomizes solutions, via recombination and mutation;
2. There are some symmetry in the solutions, which we do not see in GA;
3. SA finds solutions that puts zeros (lowest vulnerability) on the floodplain; and
4. SA took approximately eight hours to complete the iteration runs.

| | | | | | |
|---|---|---|---|---|---|
| 11 | 11 | 10 | 11 | 1 | 0 |
| 11 | 11 | 10 | 10 | 0 | 1 |
| 11 | 10 | 1 | 0 | 1 | 11 |
| 11 | 1 | 0 | 10 | 11 | 11 |
| 1 | 0 | 1 | 11 | 11 | 10 |
| 0 | 1 | 11 | 11 | 11 | 10 |

**Urbanization:** This design keeps the floodplain free from urbanization.

| | | | | | |
|---|---|---|---|---|---|
| 1 | 11 | 11 | 11 | 0 | 0 |
| 1 | 0 | 1 | 1 | 0 | 0 |
| 10 | 10 | 1 | 0 | 0 | 10 |
| 10 | 0 | 0 | 0 | 11 | 10 |
| 0 | 0 | 0 | 11 | 0 | 11 |
| 0 | 0 | 11 | 10 | 1 | 10 |

**Literacy:** It avoided putting ‘illiterate’ areas, 11, on or near the floodplain.

| | | | | | |
|---|---|---|---|---|---|
| 10 | 10 | 10 | 10 | 1 | 0 |
| 10 | 10 | 11 | 1 | 0 | 1 |
| 10 | 10 | 1 | 0 | 1 | 1 |
| 10 | 0 | 0 | 1 | 10 | 10 |
| 1 | 0 | 1 | 10 | 10 | 10 |
| 0 | 0 | 10 | 10 | 10 | 10 |

**Mortality:** This places lowest mortality areas on the floodplain.

| | | | | | |
|---|---|---|---|---|---|
| 10 | 10 | 10 | 10 | 0 | 0 |
| 10 | 1 | 10 | 1 | 0 | 1 |
| 10 | 10 | 1 | 0 | 1 | 10 |
| 10 | 1 | 0 | 1 | 10 | 10 |
| 1 | 0 | 1 | 10 | 10 | 11 |
| 0 | 1 | 10 | 10 | 10 | 10 |

**Poverty (incidence):** The rich areas, 0, were situated on the flood plain; only one poor area was allowed in this city.

| | | | | | |
|---|---|---|---|---|---|
| 1 | 11 | 11 | 11 | 0 | 0 |
| 11 | 11 | 11 | 0 | 0 | 1 |
| 10 | 1 | 0 | 0 | 0 | 10 |
| 11 | 0 | 0 | 0 | 11 | 10 |
| 0 | 0 | 0 | 0 | 10 | 11 |
| 0 | 0 | 10 | 11 | 11 | 11 |

**Radio/ TV Penetration:** The most-informed areas, 0, were placed on floodplain.

| | | | | | |
|---|---|---|---|---|---|
| 11 | 11 | 11 | 11 | 1 | 0 |
| 10 | 11 | 11 | 0 | 0 | 0 |
| 0 | 11 | 0 | 0 | 0 | 1 |
| 11 | 0 | 0 | 0 | 10 | 10 |
| 0 | 0 | 1 | 10 | 11 | 11 |
| 0 | 0 | 0 | 11 | 11 | 0 |

**State of non-structural measures:** Only totally compliant areas, 0,are situated in the floodplain.

| | | | | | |
|---|---|---|---|---|---|
| 11 | 10 | 11 | 11 | 1 | 0 |
| 10 | 10 | 10 | 1 | 0 | 1 |
| 11 | 11 | 10 | 0 | 1 | 11 |
| 10 | 1 | 0 | 1 | 10 | 11 |
| 10 | 0 | 1 | 11 | 10 | 11 |
| 0 | 1 | 11 | 11 | 10 | 11 |

**State of structural measures:** Managed to assign the most protected areas, 0, on the floodplain.

Although Simulated Annealing produced technically better results, it seems almost unrealistic to implement in real life, e.g., SA handles the poverty chromosome by turning the floodplain into a “millionaire’s row”, and very unrealistically allows only one very poor area in the city of 36 barangays;

The SA result gives us the insight that, on the floodplain, we could strive for strict enforcement of laws that protect from flooding while GA tells us to avoid putting high-risk areas (relative to any given chromosome) right on the flood plain, e.g., it tries to avoid situating very poor areas on the flood plain, and places worse offenders of laws away from the floodplain.

## SUGGESTIONS FOR FURTHER STUDY

The model and its recommendations are dependent on the previous assumptions made on vulnerability and penalty equations, and both may still be developed and improved in further studies.

Einarsson and Rausand attempted to prioritize the components of vulnerability for industrial systems in [8] while the state government of Michigan, USA, attempted to prioritize the hazards as published [9].

There is an ongoing experiment adopting the algorithm in [9] to prioritize components of vulnerability as well as allowing non-liner interactions between related components on poverty with mortality, and literacy with radio/TV penetration.

The penalty equations may be improved by interpolating available data on the components of vulnerability.

## REFERENCES

[1] Hyogo Framework for Action. UN World Conference on Disaster Reduction, Kobe, Japan. 2005. http://www.unisdr.org/eng/hfa/htm.

[2] Kumar, T (2008). T.J. Watson Research Center, IBM Research, Yorktown Heights, New York. http://www.watson.ibm.com.

[3] Balling, R.J., Taber, J.T., Brown, M.R. and Day, K. (1999). Multiobjective Urban Planning Using Genetic Algorithm. Journal of Urban Planning and Development, 86-99.

[4] Banting, J.M, Olaes, A.M., Boñgolan, V.P., Aquino, C.R. and Ballesteros, F.C. A Genetic Algorithms Approach to Flood Disaster Management and Risk Assessment. Proceedings of the 8th National Conference on Information Technology Education (NCITE), Aklan, Philippines (2010).

[5] Olaes, A.M., Banting, J.M., Boñgolan, V.P., Aquino, C.R. and Ballesteros, F.C. Simulated Annealing In Optimizing Flooding Vulnerability and Cost. Proceedings of 3rd Asean Civil and 3rd Asean Environmental Engineering Conferences, Manila, (2010).

[6] UN/ISDR. (2004). Living with Risk: A Global Review of Disaster Reduction Initiatives. Geneva, Switzerland: United Nations.

[7] Lo, D.S. & Oreta, W.C. (2010). Seismic Risk Mapping at Micro-Scale: the Case of Barangay Carmen, Cagayan de Oro City, Philippines. Conference on “Harnessing Lessons Towards an Earthquake-Resilient Nation”. July 15-16, 2010. PhiVolcs, Quezon City.

[8] Einarsson, S. and Rausand, M. (1998) An Approach to Vulnerability Analysis of Complex Industrial Systems. Risk Analysis, Vol. 18, No. 5, 1998

[9] www.michigan.gov/documents/7pub207_60741_7.pdf [accessed 6 April 2011]